\documentclass[10pt,twocolumn,letterpaper]{article}

\usepackage{cvpr}              

\usepackage[dvipsnames]{xcolor}

\definecolor{cvprblue}{rgb}{0.21,0.49,0.74}
\usepackage[pagebackref,breaklinks,colorlinks,citecolor=cvprblue]{hyperref}
\usepackage{algorithm}
\usepackage{algorithmic}
\usepackage{flushend}

\def\paperID{5921} 
\def\confName{CVPR}
\def\confYear{2024}

\title{Uncovering the Text Embedding in Text-to-Image Diffusion Models}

\author{
    Hu Yu$^{1}$ \quad \quad \quad Hao Luo$^{2}$ \quad \quad \quad
    Fan Wang$^{2}$ \quad \quad \quad Feng Zhao $^{1}$ \\	  
    $^1$University of Science and Technology of China \quad  \quad	$^2$Alibaba Group\\
    \texttt{yuhu520@mail.ustc.edu.cn} \quad  \texttt{fzhao956@ustc.edu.cn}\\
}

\begin{document}

\let\oldtwocolumn\twocolumn
\renewcommand\twocolumn[1][]{
    \oldtwocolumn[{#1}{
    \begin{center}
    \includegraphics[width=\textwidth]{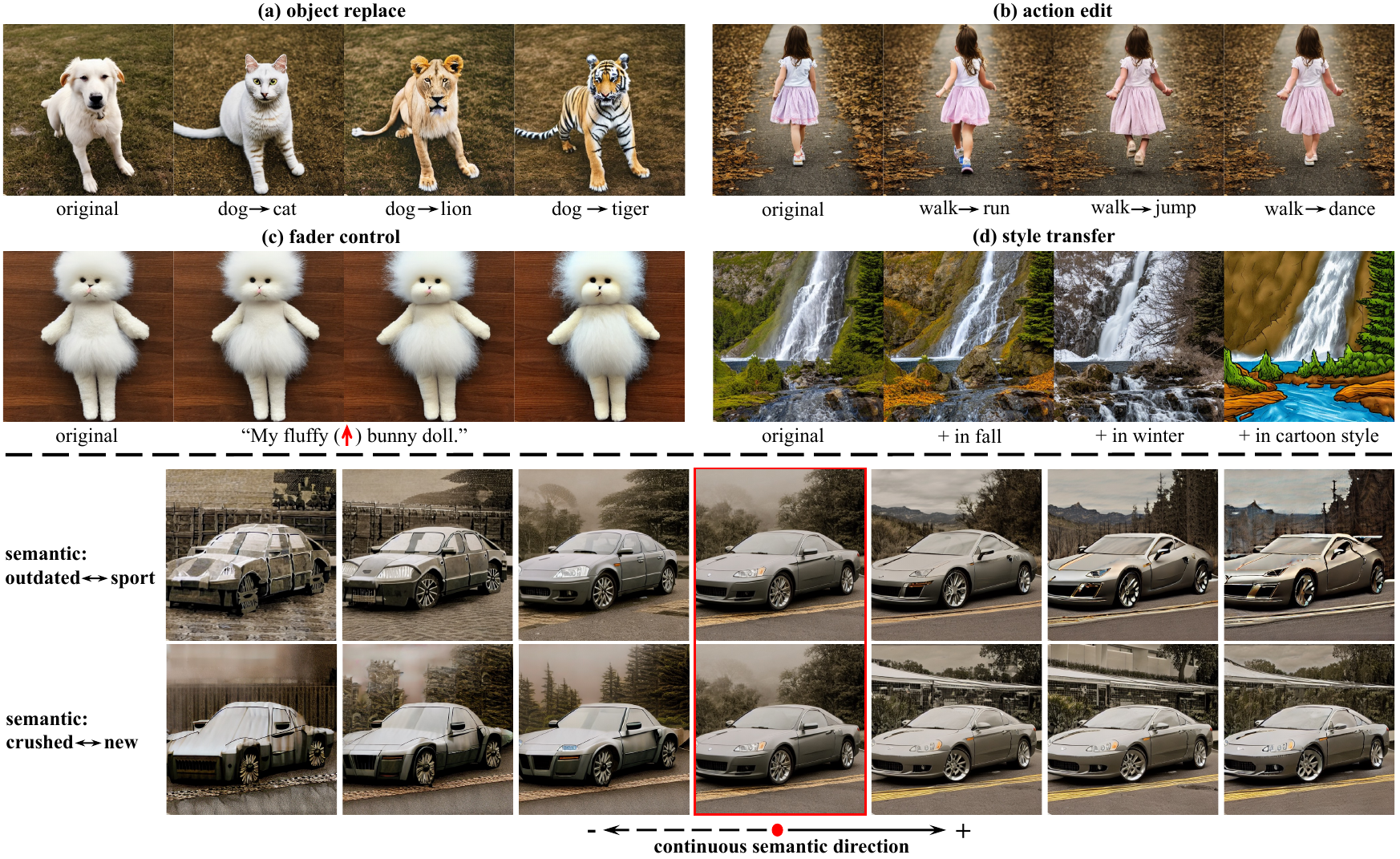}
    \setlength{\abovecaptionskip}{-0.4cm}
    \setlength{\belowcaptionskip}{-0.2cm}
    \captionof{figure}{\textbf{Examples of controllable image editing and explicable semantic directions disentangled from the text embedding of the stable diffusion model.} For controllable image editing (upper part), our method can achieve local-level object replacement, action editing, fader control and global-level style transfer via simply manipulating the text embedding. For explicable semantic directions (bottom part), we discover that the singular vector of the singular value decomposition (SVD) of the text embedding is inherent semantic space, with different singular vectors represent different semantic directions. All these are reached in a learning-free manner.}
    \label{fig:mainapp}
    \end{center}
    }]
}

\maketitle
\begin{abstract}
\vspace{-0.4cm}

The correspondence between input text and the generated image exhibits opacity, wherein minor textual modifications can induce substantial deviations in the generated image. While, text embedding, as the pivotal intermediary between text and images, remains relatively underexplored. In this paper, we address this research gap by delving into the text embedding space, unleashing its capacity for controllable image editing and explicable semantic direction attributes within a learning-free framework. Specifically, we identify two critical insights regarding the importance of per-word embedding and their contextual correlations within text embedding, providing instructive principles for learning-free image editing. Additionally, we find that text embedding inherently possesses diverse semantic potentials, and further reveal this property through the lens of singular value decomposition (SVD). These uncovered properties offer practical utility for image editing and semantic discovery. More importantly, we expect the in-depth analyses and findings of the text embedding can enhance the understanding of text-to-image diffusion models.

\end{abstract}


\section{Introduction}
\label{sec:intro}

Text-to-image diffusion models have gained remarkable popularity and demonstrated significant capabilities in image generation~\citep{rombach2022high,nichol2021glide,ramesh2022hierarchical,yu2022scaling,saharia2022photorealistic}. These models empower the generation of captivating images portraying diverse scenes and styles based on user-provided free text descriptions, finding widespread applications in various domains, including artistic creation and medical imaging\cite{chambon2022adapting}. However, the mapping between input text and the generated image exhibits opacity, wherein minor textual modifications can induce substantial deviations in the generated image. To illustrate, consider a scenario where a user generates an image featuring a dog and expresses satisfaction with the overall result. Subsequently, the user wants to replace the dog with a cat while preserving the remaining elements. Achieving this modification solely through a text prompt adjustment proves challenging and intractable \cite{hertz2022prompt}. Illustrative examples are presented in Figure \ref{fig:manual}(a).

Recent works \cite{gal2022image, kawar2023imagic, ruiz2023dreambooth} find that text embedding, as the bridge between texts and images, is a suitable learning space for achieving controllable edits. Although these works make great attempts to reveal the rationality and feasibility of treating text embedding as an editing space, they predominantly adopt a learning-based manner that requires customized training to learn the appropriate text embedding at each editing time. Besides, due to neglect to comprehensively analyse the inherent property of text embedding, they also did not explore its potential in learning-free editing.

In this paper, we aim to fill this gap with comprehensive exploration of text embedding space in stable diffusion, unveiling the latent potential. Our investigation reveals two principal applications: controllable image editing ability and an explicable semantic direction attribute.

The property investigation of text embedding begins by examining two transition mappings. Firstly, we scrutinize the text encoder to investigate the transition from text to text embedding, identifying two attention mask mechanisms that exert significant influence on the transition map. This introduces the first insight about the context correlation within text embedding. Concisely, the causal mask ensures that specific word embedding is solely correlated with the word embeddings preceding it. The absence of the padding mask endows the padding embedding with information from the semantic embedding.
Subsequently, we explore the transition from text embedding to the generated image using a ``mask-then-generate" strategy. With this way, we get the second insight about the meaning and significance of per-word embedding. Concisely, the absence of a single word embedding does not alter the overall content. Semantic embeddings weight more than padding embeddings, and they can achieve content and style disentanglement. Refer to Subsec. \ref{subsec:insight} for details of these two insights.
Moreover, we introduce the applications of the obtained properties. Under the recovered principled guidance, we can achieve controllable image editing, including object and action replacement, fader control, and style transfer, with simple text embedding modifications, as depicted in upper part of Fig.~\ref{fig:mainapp}.
Additionally, we employ an optimization paradigm to further underpin the rationale of learning-free editing operations. 


Besides the properties concealed within the bridging role of text embedding in text-to-image diffusion model, our attention naturally transitions to another intrinsic characteristic of text embedding, semantic diversity. We recognize that natural language descriptions are highly abstract and semantically diverse, yielding numerous images matching a fixed text. Our exploration draws inspiration from this observation to reveal a similar potential in text-to-image diffusion models. We ascertain that text embedding inherently possesses diverse semantic potential, enabling the generation of distinct images with varying semantics even under a fixed seed and text. This additional semantic diversity is uncovered through the lens of singular value decomposition (SVD) \cite{golub1980analysis}. Our findings demonstrate that the singular vectors of the SVD of text embedding delineate a desirable semantic space, as illustrated in the bottom part of Fig.~\ref{fig:mainapp}.

To demonstrate the generality and efficacy of our approach, we conduct extensive experiments using a diverse range of images from various scenarios. Furthermore, we present a qualitative comparison with some representative and top-tier learning-free image editing methods \cite{hertz2022prompt, wu2023uncovering}, wherein our method exhibits comparable or superior performance in a straightforward and unexpected manner. 

\section{Related Works}
\label{sec:related}

\subsection{Text Embedding}
Text embedding plays a fundamental and indispensable role in text-to-image diffusion models, serving as the vital link between input text and output image. Previous research has demonstrated the expressive capability of this embedding space in capturing basic image semantics \cite{tsimpoukelli2021multimodal,cohen2022my}. Several methodologies have explored the potential value of the text embedding space across various applications \cite{gal2022image, kawar2023imagic, ruiz2023dreambooth, wu2023uncovering}. For instance, the approach of textual inversion \cite{gal2022image} involves learning to transform image concepts into new text embeddings using a frozen text-to-image model. It then combines this new text embedding with other textual descriptions to achieve personalized generation. Imagic \cite{kawar2023imagic}, on the other hand, learns a text embedding aligned with both the input image and the target text while fine-tuning the diffusion model to capture image-specific appearances.
These methods demonstrate impressive performance utilizing text embedding; however, they necessitate training at each editing instance and overlook an exploration of the inherent properties of text embedding.

\subsection{Tuning-Free Image Editing}
Controllable image editing poses a critical and formidable challenge in diffusion models. Numerous methods have demonstrated impressive performance in image generation and editing \cite{ruiz2023dreambooth, zhang2023adding, kim2022diffusionclip, yang2023paint, mokady2023null}. However, a majority of these approaches necessitate architectural modifications and model fine-tuning. In contrast, tuning-free image editing methods \cite{hertz2022prompt,tumanyan2023plug,meng2021sdedit, avrahami2022blended, wu2023uncovering, couairon2022diffedit} offer enhanced flexibility and efficiency, albeit requiring a thorough analysis and understanding of the opaque generation process. For instance, some tuning-free image editing methodologies leverage the correlation between cross-attention and structure \cite{hertz2022prompt,tumanyan2023plug}. Wu et al. \cite{wu2023uncovering} proposed to first employ the original text embedding in the early step to generate content and then switch to new text embedding in the latter step to achieve image editing. In this paper, we advance tuning-free image editing from the perspective of text embedding.

\subsection{Semantic Space in Generative Models}
Semantic property has garnered attention in the context of Generative Adversarial Networks (GANs) \cite{harkonen2020ganspace,shen2020interpreting,shen2021closed}, where various factorization techniques are employed to define meaningful directions. Notably, GANSpace \cite{harkonen2020ganspace} identifies interpretable semantic directions through applying principal component analysis (PCA) \cite{pearson1901liii} to specific layers of the generator. In contrast to GANs, the semantic space is less explored in diffusion models. Recently, some works find that the bottleneck feature of the UNet of the pretrained diffusion models may be a semantic latent space \cite{kwon2022diffusion,haas2023discovering,park2023unsupervised}. While these investigations make commendable strides in discovering the semantic space within the UNet of diffusion models, our approach distinguishes itself by exploring the semantic space concealed within text embedding, providing a novel perspective in comparison to existing methods.


\section{Controllable Image Editing}   \label{sec:controllable}
\noindent \setlength{\parindent}{2em}\textbf{Task description.} 
Given a source text $T^s$ and a target $T^t$, we can get their corresponding text embedding pair $e^s \in R^{L \times D}$ and $e^t \in R^{L \times D} $, and the generated image pair $\mathcal{I}^s=\mathcal{G_{DM}}(e^s)$ and $\mathcal{I}^t=\mathcal{G_{DM}}(e^t)$, where $L$ is the max text length, $D$ is the feature dimension of per word embedding, and $\mathcal{G_{DM}}$ denotes the text-to-image diffusion model.
Now, we desire to generate an image $\mathcal{I}^{*}$, which is of the background and structure of source image $\mathcal{I}^s$, but aligns with the target text $T^t$. In this paper, we achieve this goal only by mixing text embedding pair $e^s$ and $e^t$ to get the new text embedding $e^{*}$, which then generates the desired image $\mathcal{I}^{*}=\mathcal{G_{DM}}(e^{*})$. Note, we denote $e^{s}_{i} \in R^{1 \times D}, i \in [0,L-1]$ as the $i\text{-}th$ word embedding in $e^{s}$.

Note that a naive way to achieve this goal is to first employ text $T^s$ in early steps to generate the structure and then turn to text $T^t$ in the latter steps to complete the necessary details that match the description of $T^t$ \cite{wu2023uncovering}. In contrast, our method takes a more intrinsic approach. We attain the same objective with fixed text embedding across all steps, necessitating a profound understanding of text embedding.

This section is structured as follows. In Subsec. \ref{subsec:insight}, we unleash the potential and properties of the text embedding to derive general insights guiding learning-free editing. In Subsec. \ref{subsec:edit}, we introduce applications of the obtained conclusions and insights, proposing specific editing operations. In Subsec. \ref{subsec:optimize}, we employ an optional optimization paradigm to substantiate the rationale behind learning-free editing operations.
Additional background knowledge of diffusion models is available in supplementary material.

\begin{figure}[h]
    \begin{center}
	\includegraphics[width=0.98\linewidth]{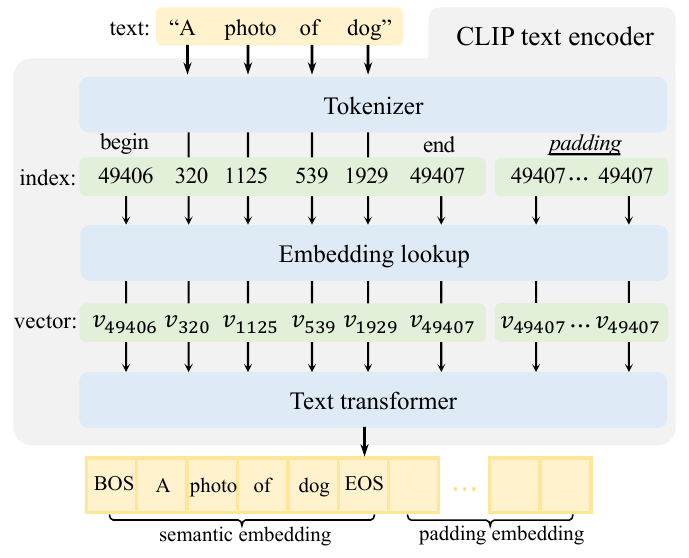}
    \end{center}
    \caption{The flow chart of the text encoder in CLIP. We take the text ``A photo of dog" as example. The given text prompt passes through the tokenizer, embedding lookup, and text transformer to get the corresponding text embedding.} 
    \label{fig:encoder}
\end{figure}

\subsection{Text Embedding} \label{subsec:insight}


Our investigation commences with an analysis of the text encoder, specifically focusing on the transition from text to text embedding. In this study, we employ the CLIP text encoder \cite{radford2021learning} as the default text encoder in text-to-image diffusion models, such as stable diffusion. The visualization of the text encoder procedure to obtain text embedding is illustrated in Fig.~\ref{fig:encoder}. Initially, each word in an input string is transformed into a token, representing an index in a pre-defined dictionary. Each token is then associated with a unique embedding vector, retrievable through an index-based lookup. These embedding vectors undergo processing by a transformer to generate the text embedding. To accommodate different text lengths, padding is applied to extend the index number to a specified maximum length $L$ after tokenizer. Consequently, the resulting text embedding can be dissected into two components: the semantic embedding and the padding embedding.

Through the aforementioned procedure, we identify two attention mask mechanisms in the text encoder that govern the context correlation within the text embedding: the causal mask and padding mask \cite{vaswani2017attention}. The CLIP text encoder employs a causal mask and omits the padding mask (details and visualization in supplementary material). This design choice imparts \textbf{the first insight}: the context correlation within the text embedding. Specifically, (1) the use of a causal mask ensures that information in a specific word embedding is solely correlated with the word embedding preceding it. (2) The absence of the padding mask endows the padding embedding with information from the semantic embedding.

\begin{figure}[h]
    \begin{center}
	\includegraphics[width=0.98\linewidth]{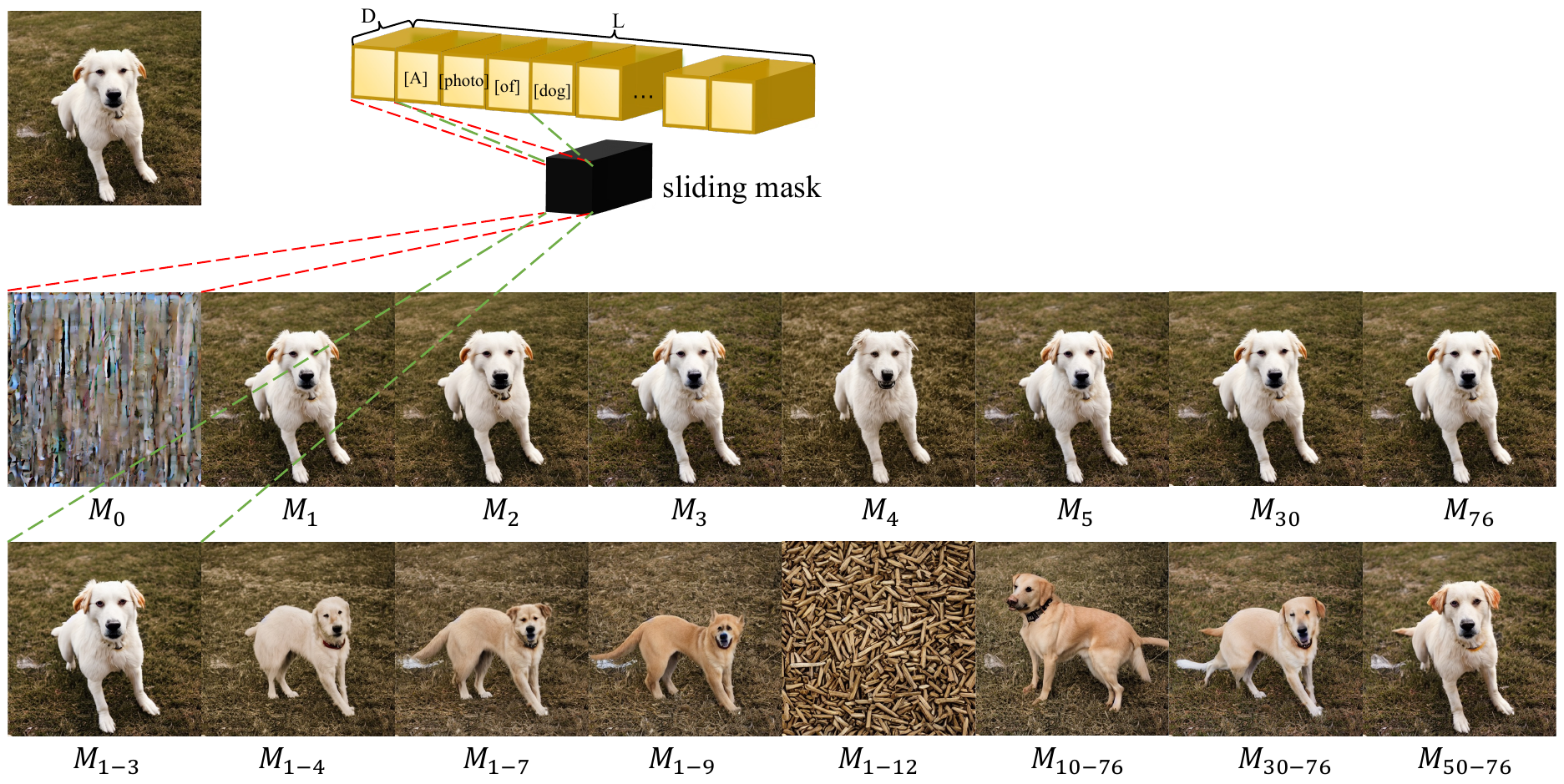}
    \end{center}
    \caption{Examples of the \textit{mask-then-generate} strategy. In this strategy, we mask certain word embeddings and compare the resulting images. For example, $M_{i\text{-}j}$ denotes the generated image with the $i\text{-}th$ to $j\text{-}th$ word embeddings masked.} 
    \label{fig:mask}
\end{figure}

Furthermore, we seek to discern the meaning and significance of per-word embedding by scrutinizing the transition from text embedding to generated image. To accomplish this, we adopt a \textit{mask-then-generate} strategy and conduct numerous experiments. In particular, we mask specific word embeddings and generate corresponding images, as illustrated in Fig.~\ref{fig:mask}. Additional analyses and experiments are available in the supplementary material. These experiments introduce \textbf{the second insight}: the meaning and significance of per-word embedding. Specifically, (1) the absence of a single word embedding does not alter the overall content, except for the BOS embedding, as depicted in the top row of Fig.~\ref{fig:mask}. The BOS embedding is semantically meaningless but indispensable for image generation in stable diffusion, as its consistency across different text embeddings has been learned during training. 
(2) The semantic embedding takes precedence over the padding embedding. For instance, blocking the semantic embedding significantly influences the generation of the original image, while blocking an equivalent amount of word embedding in the padding embedding has negligible impact. Within the semantic embedding, the embedding of meaningful words (e.g., objects, descriptive words, or action words) hold greater importance than others. Within the padding embedding, the importance decreases as the distance to the semantic embedding increases.
(3) Content and style disentanglement can be achieved through the semantic and padding embeddings. The semantic embedding encapsulates the majority of information in the text embedding, representing image content, while the padding embedding carries less information and can be viewed as representing image style.

\begin{figure*}[t]
	\begin{center}
		\includegraphics[width=0.98\linewidth]{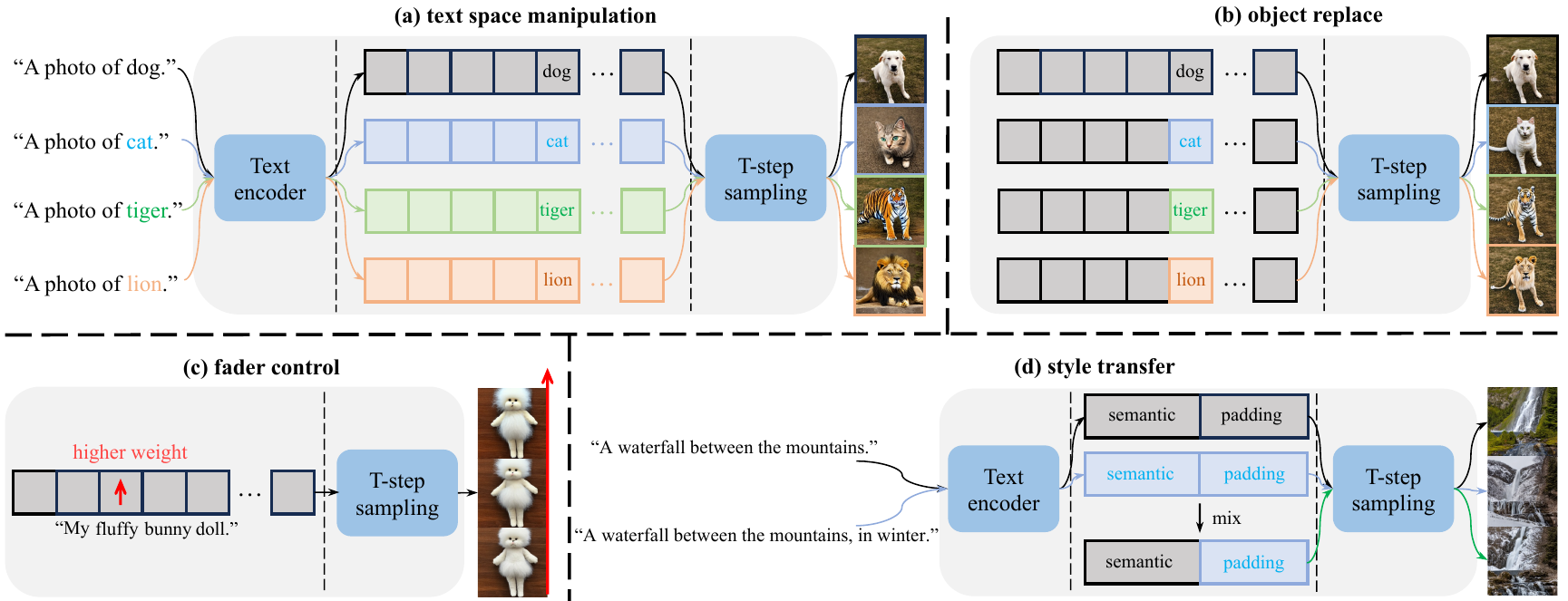}
	\end{center}
	\caption{Manipulation in text space and text embedding space. (a) Replacing in the text space leads to random and uncontrollable image content. Slight change of text induces significant and uncontrollable deviation in the generated image. (b) Controllable object replacement is achievable by substituting key word embeddings in the text embedding space. (c) Re-scaling the weight of the descriptive word embedding leads to continuous fader control. (d) Style transfer is possible via disentangling the content and style in text embedding.}
	\label{fig:manual}
\end{figure*}

\subsection{Learning-free Editing}  \label{subsec:edit}
Building upon these insights and observations, we delve into the potential avenues for accomplishing learning-free image editing through straightforward text embedding manipulation. Concretely, we start by proposing a general framework, followed by the details of the specific editing operations. Formally, our general algorithm is:
\begin{algorithm}[h]
\caption{Text embedding based image editing.}
\label{alg:training}
\begin{algorithmic}[1]
    \REQUIRE Source text $T^s$ and Target text $T^t$.
    \STATE $e^s = Text \ encoder(T^s)$ 
    \STATE $e^t = Text \ encoder(T^t)$    
    \STATE $e^{*} = mix(e^s, e^t)$ 
    \STATE $\mathcal{I}^s = \mathcal{G_{DM}}(e^s)$
    \STATE $\mathcal{I}^{*} = \mathcal{G_{DM}}(e^{*})$
    \RETURN ($\mathcal{I}_s$, $\mathcal{I}^{*}$)
\end{algorithmic}
\end{algorithm}

In light of this general formulation, we proceed to delineate the mixing function $mix(e^s, e^t)$ within various specific editing operations, as illustrated in Fig.~\ref{fig:manual}.

\noindent \setlength{\parindent}{2em}\textbf{Single Word Embedding Swap.}
Replacement is a common type of editing operation. For instance, a user may change the text from ``a photo of a dog" to ``a photo of a cat," intending solely to alter the dog to a cat while maintaining the background. We address this with the second insight. Given that the absence of single word embedding preserves the background and the meaningful word embedding plays a key role, we propose to only replace the specific word embedding, with the corresponding mixed  $e^{*}$ as follows:

\begin{equation}
    \small
    \label{eq:replace}
    e^{*}_i:= \begin{cases}
     e^t_i & \text{ if } T^s_i \ne T^t_i \\
     e^s_i & \text{ otherwise. }
    \end{cases}
\end{equation}
We visualize this editing type in Fig.~\ref{fig:manual} (b). Note that the replacement word can be object or action.

\noindent \setlength{\parindent}{2em}\textbf{Weight Scaling.} 
Besides, the user may wish to strengthen or weaken the extent to which certain word is affecting the resulting image. For instance, given the text ``My fluffy bunny doll," there may be a wish to control the fluffiness of the doll. To accommodate this editing type, we propose to scale the weight associated with the word embedding of the descriptive word ``fluffy." The corresponding mixed  $e^{*}$ is as follows:

\begin{equation}
    \small
    \label{eq:scale}
    e^{*}_i:= \begin{cases}
     c \cdot e^s_i & \text{ if } i = j  \\
     e^s_i & \text{ otherwise, }
    \end{cases}
\end{equation}
where $j$ is the index of the word we aim to control (e.g., for text ``My fluffy bunny doll", $j=2$ is the index of word ``fluffy" ).
Here, $c$ is the controlling weight and the original weights of all word embeddings equal to one. 
We visualize this editing type in Fig.~\ref{fig:manual} (c).

\noindent \setlength{\parindent}{2em}\textbf{Semantic and Padding Embedding Swap.}
Given the content and style disentanglement property in the second insight, it is natural to apply it to style transfer. The corresponding mixed  $e^{*}$ is as follows:
\begin{equation}
    \small
    \label{eq:style}
    e^{*}_i:= \begin{cases}
     e^s_i & \text{ if } e^s_i \in  semantic \ part \\
     e^t_i & \text{ if } e^t_i \in  padding \ part.
    \end{cases}
\end{equation}
We visualize this editing type in Fig.~\ref{fig:manual} (d).

\subsection{Optimizing-based Paradigm}   \label{subsec:optimize}
\begin{figure}[ht]
	\begin{center}
		\includegraphics[width=0.98\linewidth]{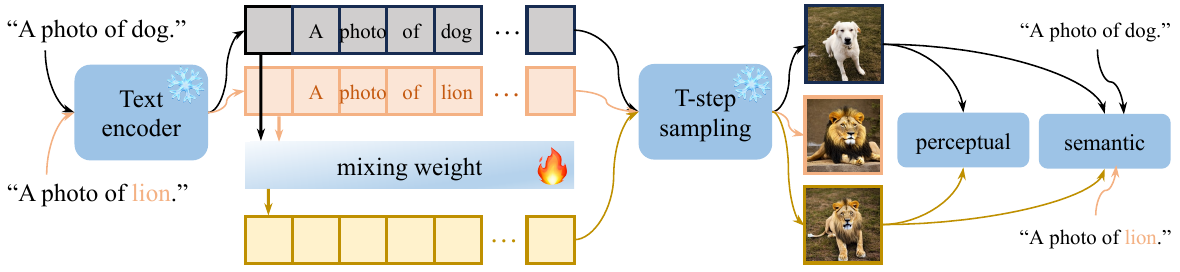}
	\end{center}
	\caption{Optional optimization framework. We freeze the parameters of the diffusion model and only learn the soft mixing weight.}
	\label{fig:learning}
\end{figure}

We employ an optimization paradigm to further support the rational of the learning-free editing operations and also as an optimal editing choice. Concretely, we propose to learn the soft mixing weight of the source and target text embedding with a widely adopted framework \cite{patashnik2021styleclip, kwon2022diffusion, wu2023uncovering} as shown in Fig.~\ref{fig:learning}. We discover that the learned soft mixing weight is highly consistent with the simple modifications, with specific cases in the Subsec. \ref{sebsec:optimizeresult}.
Concretely, the soft mixing manner is as follows:
\begin{equation}
    \small
    \label{eq:mix}
    \begin{aligned}
        e^{*} = \mathbf{\lambda} \odot e^s + (\pmb{1}-\mathbf{\lambda}) \odot e^t,\\
    \end{aligned}
\end{equation}
where $\mathbf{\lambda} \in R^{L \times 1}$ is the learnable mixing weight vector. The adopted optimization paradigm is as follows:
\begin{equation}
    \small
    \label{eq:loss}
    \begin{aligned}
        L = L_{clip}(\mathcal{I}^s, \mathcal{I}^{*}, T^s, T^t) + L_{prec}(\mathcal{I}^s, \mathcal{I}^{*}),\\
    \end{aligned}
\end{equation}
where $L_{clip}$ is the CLIP loss \cite{gal2022stylegan} for semantic match. $L_{perc}$ is the perceptual loss \cite{johnson2016perceptual, simonyan2014very} to prevents drastic content changes. Note that our method only needs to learn the combination weight $\mathbf{\lambda}$, which is as few as $L=77$ parameters. More implementation details are depicted in the supplementary material.

\subsection{Extension to Real Image Editing}
There exists slight difference between image editing setting and our task description. 
In image editing, a real image is externally given instead of first employing the text to generate such image and then performing editing. 
While, our text-guided image editing can be easily extended to real image editing with the $\mathit{inversion}$ method \cite{song2020denoising, mokady2023null, hertz2022prompt, kim2022diffusionclip, wu2023uncovering}. We depict real image editing examples in Subsec. \ref{sebsec:free}.

\section{Explicable Semantic Latent Directions} \label{sec:explicable}

In addition to the properties elucidated within the bridging role of text embedding in the text-to-image diffusion model in Sec. \ref{sec:controllable}, our focus naturally shifts to another intrinsic characteristic of text embedding: semantic diversity. 
The organization of this section is as follows. In Subsec. \ref{sebsec:motivation}, we present the motivation derived from the potential semantic diversity within a single text. Subsequently, in Subsec. \ref{sebsec:SVD}, we identify the explicable semantic direction of the text embedding space through the lens of SVD.

\subsection{Potential Semantic Diversity within Single Text} \label{sebsec:motivation}
Natural language description is highly abstract and semantically diverse. For example, given a fixed neutral text ``a photo of car", there are massive images with different semantics that match this description (e.g., a sports car, an outdated car, a new car, or a crushed car). Drawing inspiration from this inherent diversity, we explore to reveal the similar potential within the text-to-image diffusion model. The diversity in image generation within the text-to-image diffusion model arises from two sources: the variation in the text prompt and the random seed. Given a fixed text prompt, we can generate numerous different images matching the text with varying random seed. While, this diversity no longer exists with fixed seed. However, we discover that, even with fixed seed and text, text embedding inherently possesses diverse semantic potentials. Concretely, we can generate images with semantic changes via adding the additional discovered semantic directions.

\subsection{Latent Semantic Directions in SVD}  \label{sebsec:SVD}
\begin{figure}[t]
    \begin{center}
    \includegraphics[width=0.98\linewidth]{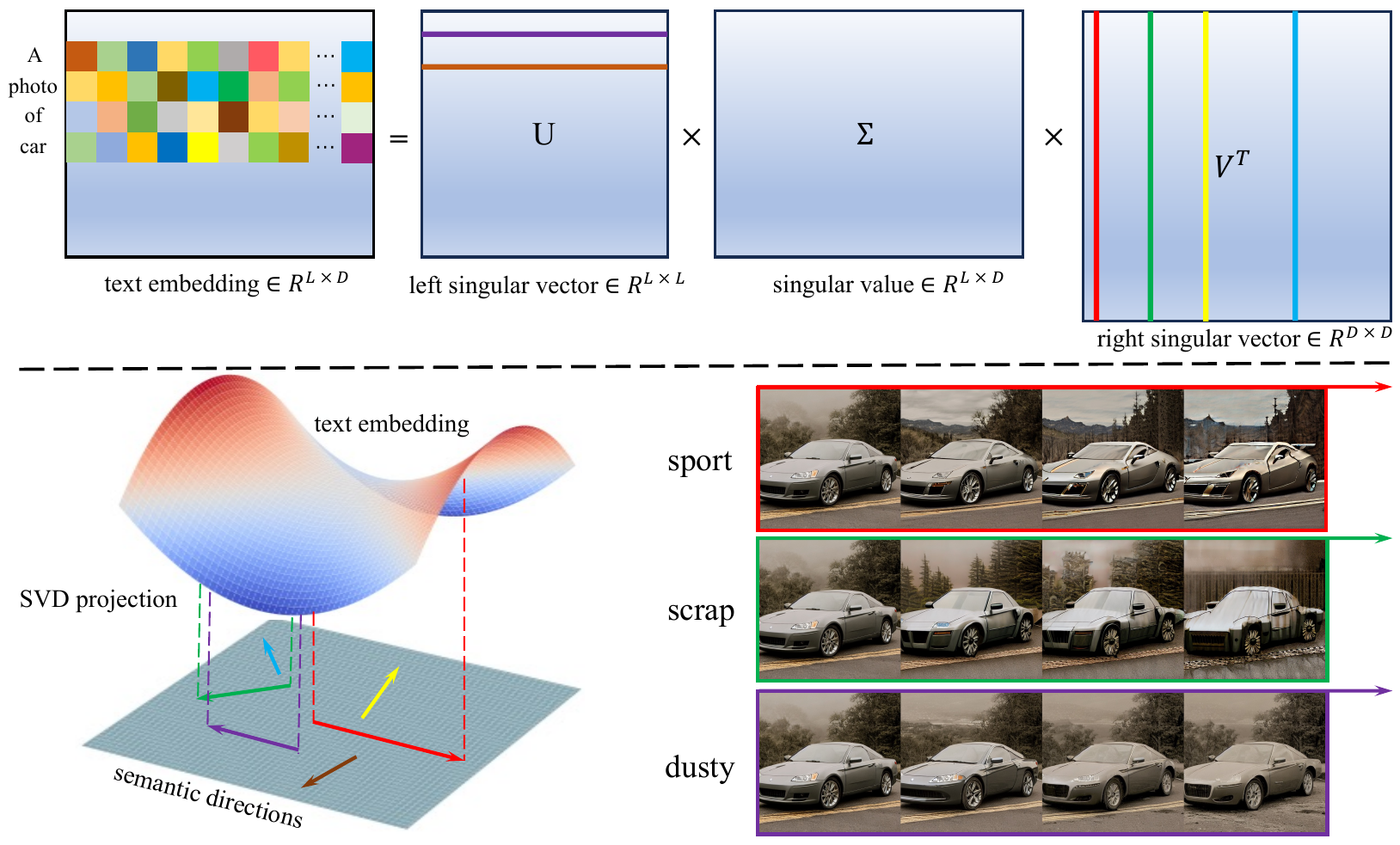}
    \end{center}
    \caption{The SVD of the text embedding matrix. The right singular vector at the column of $V^T$ and the left singular vector at the row of $U$ are desirable semantic directions, with different singular vectors represent different semantic directions.}
    \label{fig:SVD}
\end{figure}

We unleash the diverse semantic potentials of the text embedding space through the lens of singular value decomposition (SVD) as illustrated in Fig.~\ref{fig:SVD}. SVD \cite{golub1980analysis} is a classical factorization technique of matrix.
Specifically, the singular value decomposition on the text embedding matrix $e \in R^{L \times D} $ can be expressed as follows:
\begin{equation}
    \small
    \label{eq:loss}
    \begin{aligned}
    e = U \Sigma V^T,\\
    \end{aligned}
\end{equation}
where $\Sigma = diag(\sigma) \in R^{L \times D}$ is the singular value matrix, with the singular values in descending order, $U \in R^{L \times L}$ is the left singular matrix, and $V^T \in R^{D \times D}$ is the right singular matrix. We find that the right singular vector $\mathbf{v} \in R^{D \times 1}$ at the column of $V^T$ and the left singular vector $\mathbf{u} \in R^{1 \times L}$ at the row of $U$ are desirable semantic directions, with different singular vectors represent different semantic directions as shown in Fig.~\ref{fig:SVD}. Formally, the new text embedding $e^{sem}$ that can generate semantically varying images can be formulated as follows:
\begin{equation}
    \small
    \label{eq:mix}
    e^{sem} = \begin{cases}
     e + \mathcal{P}( e \cdot \mathbf{v} )  & \text{ for } right \  vector \ \mathbf{v} \\
     e + \mathcal{P}( \mathbf{u} \cdot e )  & \text{ for } left \  vector \ \mathbf{u},
    \end{cases}
\end{equation}
where $\mathcal{P}$ is the expand operation to align the dimension to that of $e$.
We further discuss the insight within this observation. A potential rational is that the singular vectors are a set of orthogonal bases, with the maximum variance and information preserved when the text embedding matrix is projected onto these bases. $(e \cdot \mathbf{v}) \in R^{L \times 1}$ compresses the text embedding to single column, and $(\mathbf{u} \cdot e) \in R^{1 \times D}$ compresses the text embedding to single row.

Note some prior methods observe that the principal component analysis (PCA) \cite{pearson1901liii} of the network features of generative models usually represents semantic directions \cite{harkonen2020ganspace,haas2023discovering}. This is not contradictory with our observation, and PCA is just special case of our conclusion. Concretely, the eigenvector in PCA corresponds to the right singular vector of SVD mathematically. Concrete derivation and discussions are available in the supplementary material.

\section{Applications}
\label{sec:applications}
After unleashing the potential of text embedding for image editing and semantic directions. In this section, we show several applications using this technique, including object replacement, action editing, fader control, style transfer, real image editing and diverse semantic directions.

\paragraph{Implementation Details}
We employ the pre-trained diffusion model $\mathrm{stable\text{-}diffusion\text{-}v1\text{-}4}$ \cite{rombach2022high} for all the experiments, where all the default hyper-parameters of the model are kept unchanged. Since our method enables learning-free image editing, the pre-trained model is frozen throughout all experiments. The generated images are of size $512 \times 512$. 


\begin{figure}[ht]
    \begin{center}
	\includegraphics[width=0.98\linewidth]{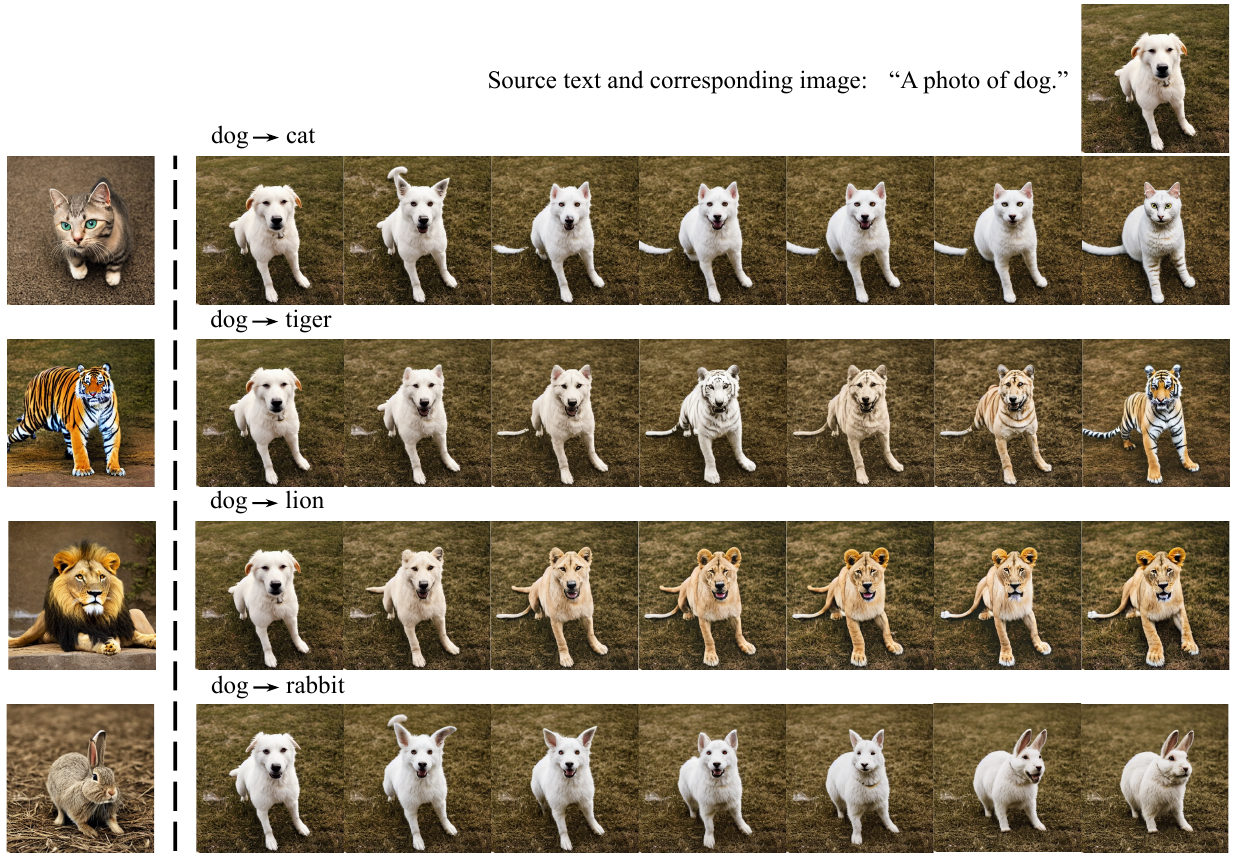}
    \end{center}
    \caption{Local editing-object replacement. On the left, we replace the word ``dog" with other words in the text space and get uncontrollable images. On the right, we replace the embedding of the word ``dog" with the embedding of other words, and achieve controllable object replacement. Note that the continuous transition is achieved by soft replacing (e.g. $w[dog] + (1-w)[cat]$ with the weight $w$ increasingly smaller from left to right).}
    \label{fig:swap}
\end{figure}

\begin{figure}[ht]
	\begin{center}
		\includegraphics[width=0.98\linewidth]{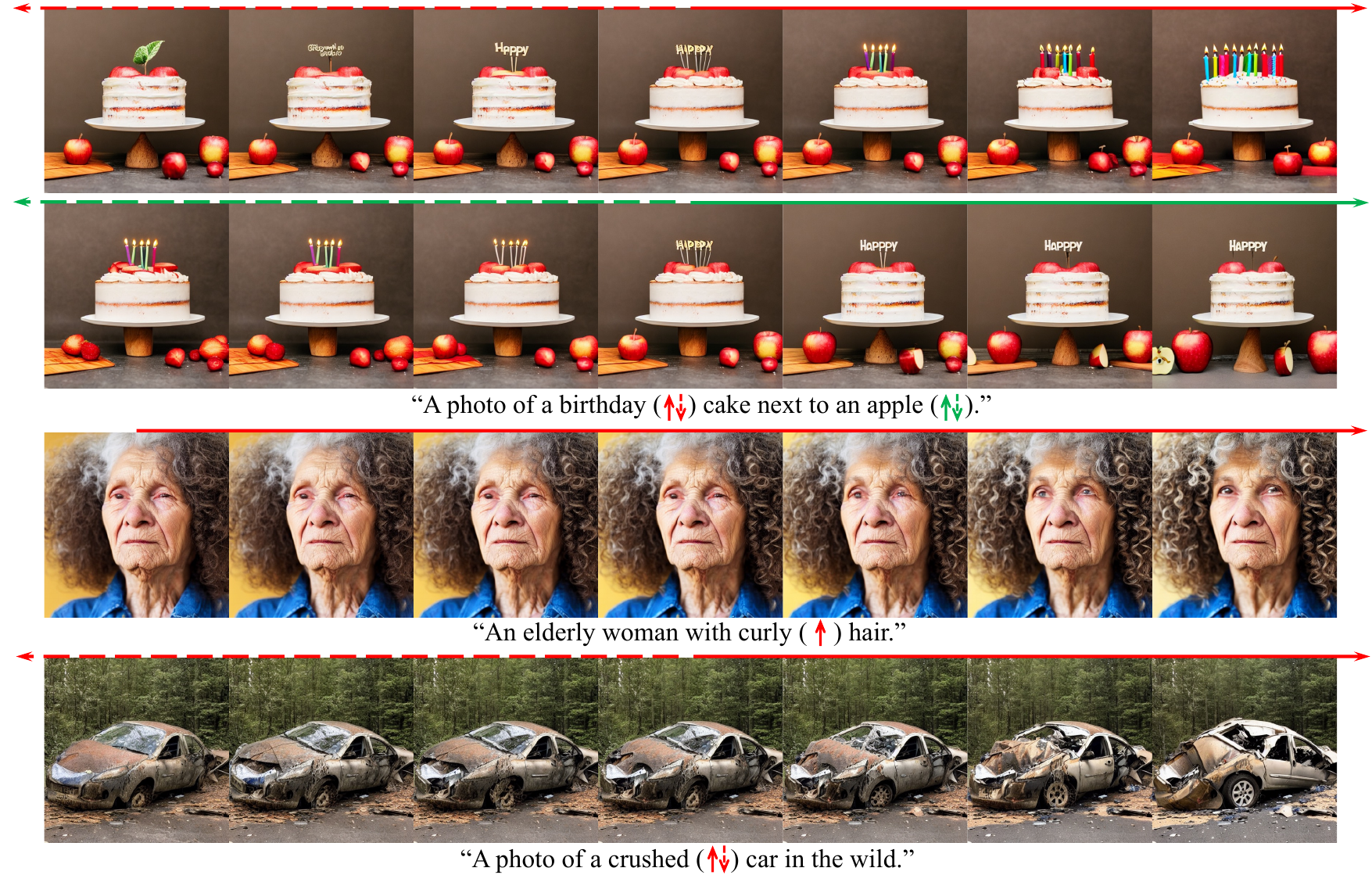}
	\end{center}
	\caption{Local editing-fader control. By reducing (dashed line) or increasing (solid line) the weight of the word embedding of the specified word (marked with an arrow), we can control the extent to which it influences the generated image.}
	\label{fig:fader}
\end{figure}

\begin{figure}[ht]
	\begin{center}
		\includegraphics[width=0.98\linewidth]{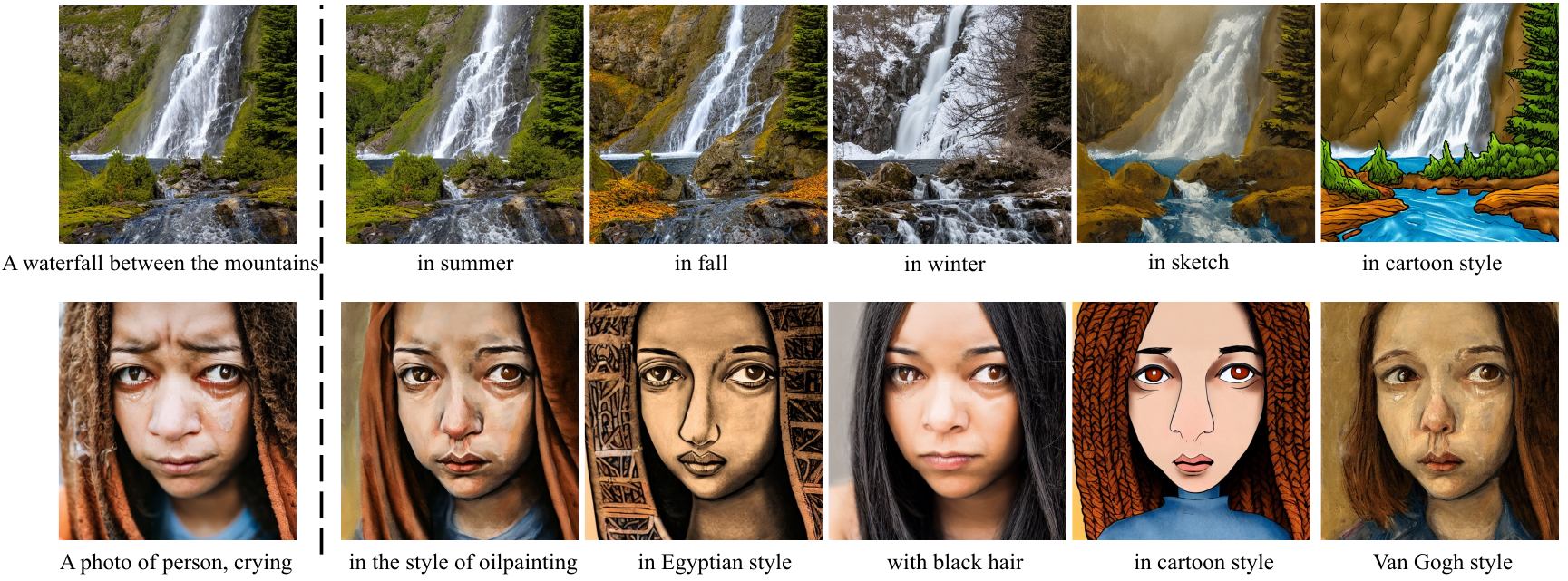}
	\end{center}
	\caption{Global editing-style transfer. By replacing the padding text embedding of the original text with that of the style description, we can create various images in the new desired styles that preserve the structure of the original image.}
	\label{fig:style}
\end{figure}

\begin{figure}[ht]
    \begin{center}
    \includegraphics[width=0.98\linewidth]{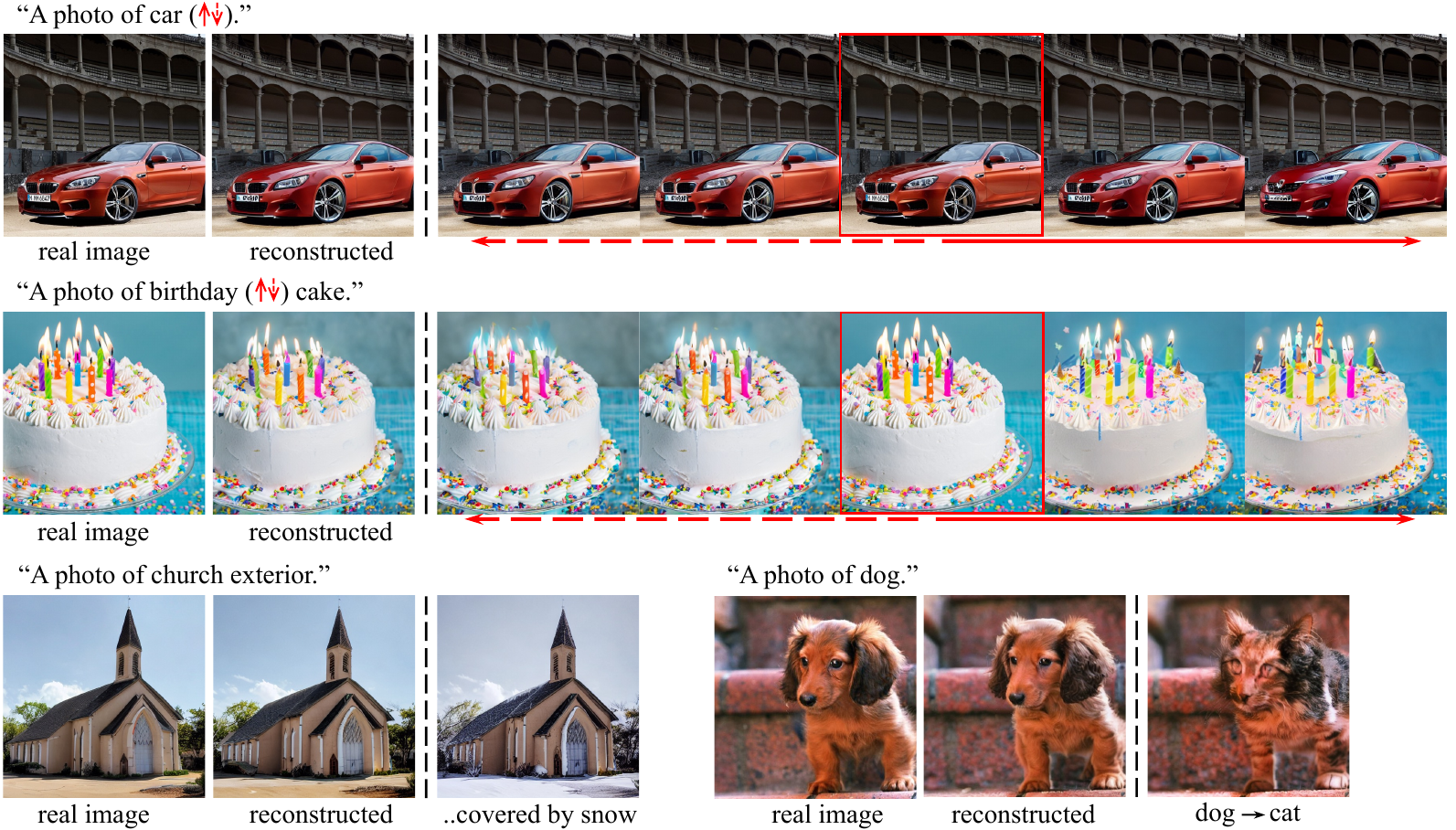}
    \end{center}
    \setlength{\abovecaptionskip}{-0.2cm}
    \setlength{\belowcaptionskip}{-0.2cm}
    \caption{Real image editing. The first column is the real image to be edited, and the second column is the inversion results using PNDM sampler \cite{liu2022pseudo}. On the right of the dashed line, we show the real image editing results of our method.}
    \label{fig:realediting}
\end{figure}

\subsection{Learning-free Image Editing}     \label{sebsec:free}
\noindent \setlength{\parindent}{2em}\textbf{Object replacement.}
We first demonstrate localized object replacement by modifying the text embedding of user-provided prompts. In Fig. \ref{fig:swap}, we depict the generated image with the user-provided prompt ``A photo of dog" (first row). Our method allows us to retain the spatial layout, geometry, and background when replacing the word ``dog" with ``cat" (right part). On the contrary, naively modifying in the text space via replacing the text to ``A photo of cat" results in a completely different geometry (left part), even when using the same random seed in the deterministic DDIM setting \cite{song2020denoising}. Our method enables smooth and continuous transition between the original object and the new object. And the replacement can be applied to diverse classes.

\noindent \setlength{\parindent}{2em}\textbf{Action editing.}
Besides object replacement, we explore the possibilities of more fine-grained control, like action editing. As shown in Fig. \ref{fig:mainapp}, given the prompt ``photo of a girl walking", we succeed to change the action ``walking" to ``running", ``jumping", and ``dancing". Besides, the body changes coherently with the corresponding action change. 

\noindent \setlength{\parindent}{2em}\textbf{Fader control.}
We can achieve image editing by replacing certain word embedding with another, while certain editing can't be reached in this way. Consider the prompt ``An elderly woman with curly hair", how to control the curvature of the hair. To achieve such editing, as depicted in Fig. \ref{fig:fader}, we perform fader control \cite{lample2017fader} to control the extent to which specified word influences the generated image by reducing (dashed line) or increasing (solid line) the weight of the word embedding of the specified word.

\noindent \setlength{\parindent}{2em}\textbf{Global Editing.}
The disentanglement of content and style in text embedding endows global editing, such as style transfer. Global editing should affect all parts of the image, but still retain the original composition, such as the location and identity of the objects. As shown in Fig. \ref{fig:style}, in the example of ``A waterfall between the mountains", we can keep the image content while changing the style to cartoon.

\noindent \setlength{\parindent}{2em}\textbf{Real Image Editing}
Following \cite{wu2023uncovering}, we employ a variant of the DDIM sampler \cite{liu2022pseudo} to conduct the $\mathit{inversion}$. The $\mathit{inversion}$ process is conditioned on the given real image and text prompt, and results in a latent noise that produces an approximation to the input image when fed to the diffusion process with the same text prompt. We depict some real image editing example in Fig.~\ref{fig:realediting}.

\subsection{Qualitative Comparison with Other Methods}    \label{sebsec:comparison}
\begin{figure}[ht]
    \begin{center}
    \includegraphics[width=0.98\linewidth]{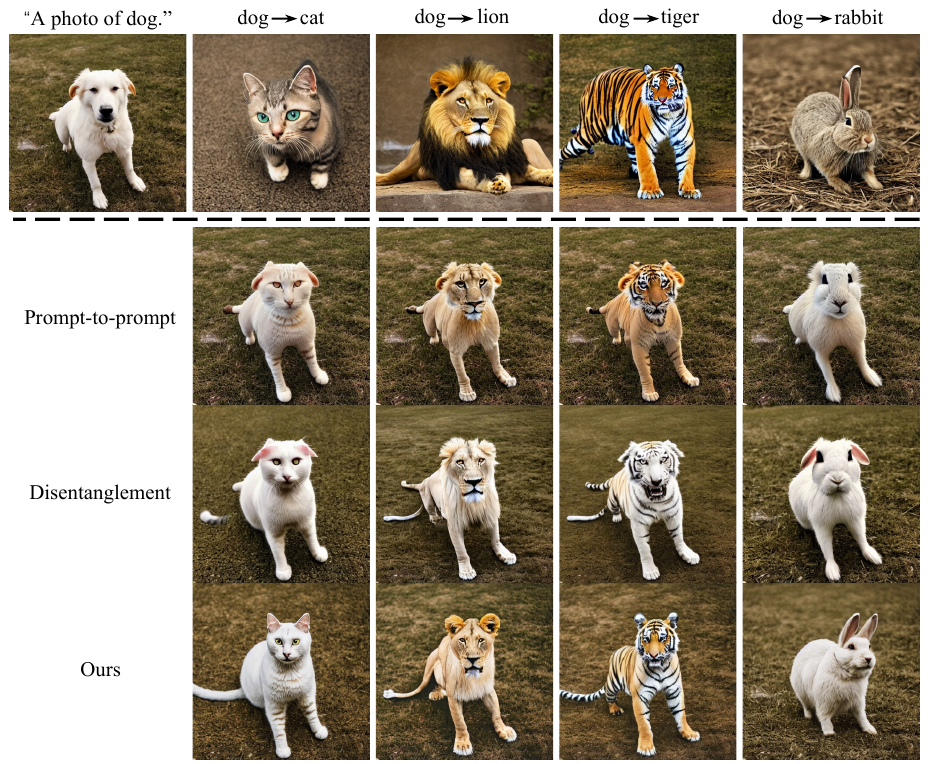}
    \end{center}
    \caption{Qualitative comparison with other learning-free image editing methods.}
    \label{fig:comparison}
\end{figure}

We qualitatively compare with two diffusion-based learning-free image editing methods, including prompt-to-prompt \cite{hertz2022prompt} and Disentanglement \cite{wu2023uncovering}, as shown in Fig.~\ref{fig:comparison}. 
Our method matches the target description better than Disentanglement, where the generated target objects are more natural.
Our method is also comparable with prompt-to-prompt with more easy operations.

\begin{figure}[ht]
    \begin{center}
    \includegraphics[width=0.98\linewidth]{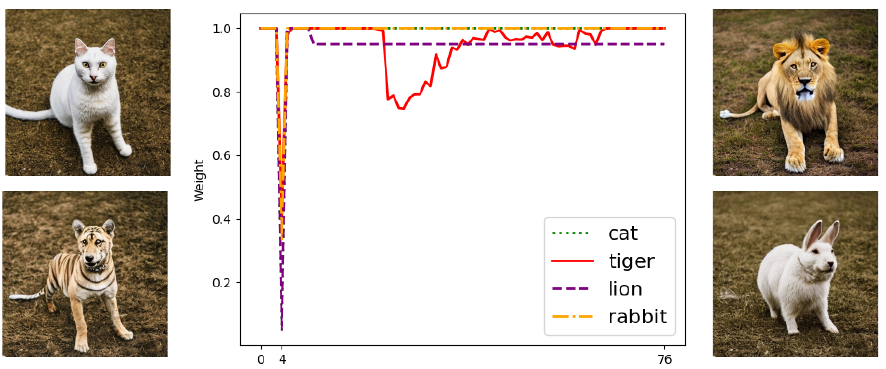}
    \end{center}
    \caption{Examples of the learned soft $\mathbf{weight}$. The images on both sides are generated with the soft mixed text embedding: $\mathbf{weight} \odot e^s + (\pmb{1}-\mathbf{weight}) \odot e^t$. The source text is ``a photo of dog", and the target text is ``a photo of cat" for cat case.}
    \label{fig:soft}
\end{figure}

\subsection{Optimization-based Image Editing}  \label{sebsec:optimizeresult}
In Fig.~\ref{fig:soft}, we present the examples generated with the optimization paradigm and the corresponding learned soft mixing weight. The examples got in learning-free manner in Fig.~\ref{fig:swap} is comparable and similar to the results reached with learning manner. And the soft mixing weight is highly consistent with Eq.~\ref{eq:replace} in learning-free mode. This optimization paradigm validates the rational and effectiveness the learning-free editing, and is also an optional editing choice.

\begin{figure}[t]
	\begin{center}
		\includegraphics[width=0.98\linewidth]{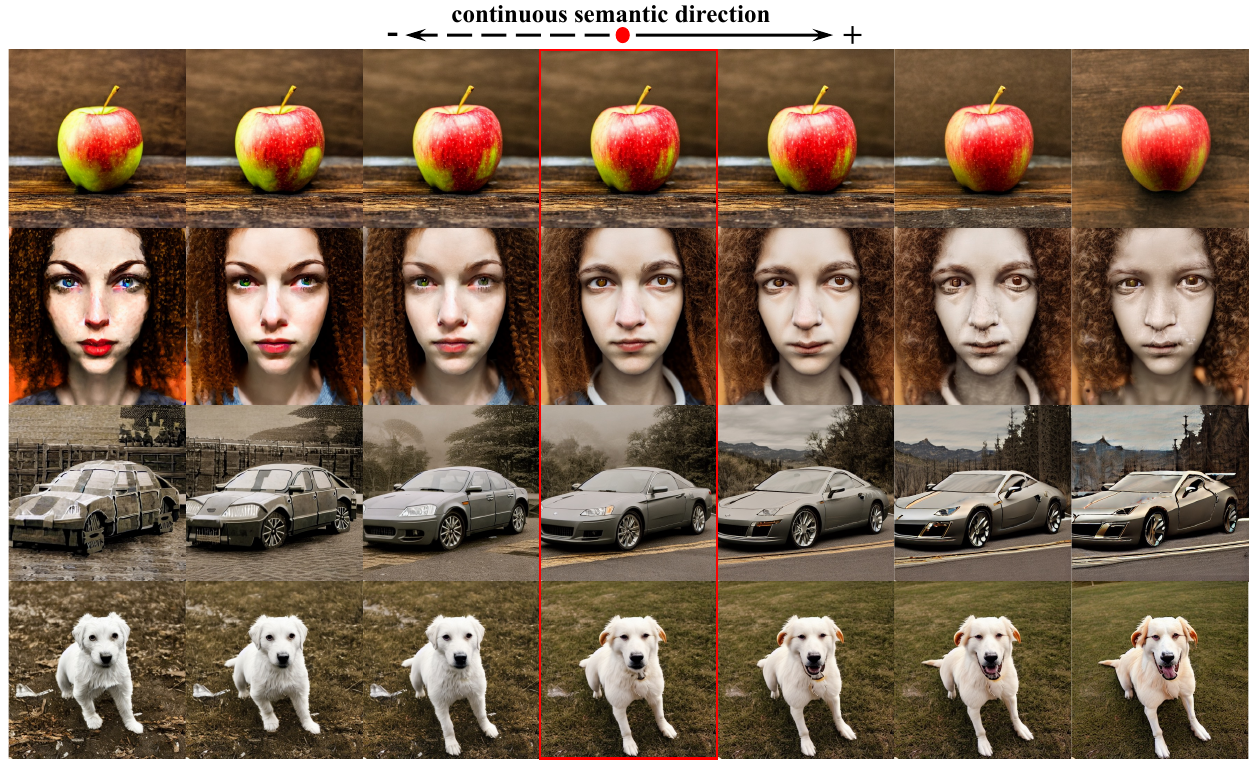}
	\end{center}
	\caption{Semantic directions. SVD in text embedding space provides a way for discovering disentangled and semantically meaningful directions. Here we show some example directions.}
	\label{fig:semantic}
\end{figure}

\subsection{Diverse Semantic Directions}
Apart from image editing capacity, another desirable attribute of text embedding is meaningful and explicable semantic directions. As shown in Fig.~\ref{fig:semantic}, we visualize some semantic directions of different text embedding. For the example of ``A photo of car", we find that the first right singular vector of its text embedding has the semantic ranging from $\mathit{outdated}$ to $\mathit{sprot}$ (third row).
Besides, such semantic space is continuous and bi-directional, enabling continuous interpolation for each semantic. We empirically find that almost all singular vectors have its semantic meaning, similar to the conclusion in GANs \cite{harkonen2020ganspace}.

\section{Conclusion}
\label{conclusion}
In this paper, we delve into the text embedding space and unleash its controllable image editing ability and explicable semantic direction attribute in a learning-free manner. We identify two key insights about the significance of per word embedding and their context correlation. Besides, we find that text embedding is naturally endowed with diverse semantic potentials. These uncovered properties can be applied to image editing and semantic discovery. These in-depth analyses and findings may contribute to the understanding of text-to-image diffusion models.

\newpage
{
    \small
    \bibliographystyle{ieeenat_fullname}
    \bibliography{main}
}

\end{document}


\maketitle

\section{Background Knowledge of Diffusion Models}  \label{sec:background}

DDPM is a latent variable model specified by a T-step Markov chain, which approximates a data distribution q(x) with a model $f(\theta)$. It contains two processes: the forward diffusion process and the reverse denoise process.

{\bf The forward diffusion process.}
The forward diffusion process starts from a clean data sample $x_0$ and repeatedly injects Gaussian noise according to the transition kernel $q(x_t|x_{t-1})$ as follows:
\begin{equation}
	\small
	\label{1}
	q(x_t|x_{t-1}) = N(x_t;\sqrt{\alpha_t}x_{t-1},(1-\alpha_t)I),  \\
\end{equation}
where $\alpha_t$ can be learned by reparameterization~\cite{kingma2013auto} or held constant as hyper-parameters, controlling the variance of noise added at each step.
From the Gaussian diffusion process, we can derivate closed-form expressions for the marginal distribution $q(x_t|x_0)$ and the reverse diffusion step $q(x_{t-1}|x_t,x_0)$ as follows:
\begin{align}
\label{2}
    q(x_t|x_0) = N(x_t;\sqrt{\bar{\alpha} _t}x_0,(1-\bar{\alpha}_t)I), \\
\label{3}
    q(x_{t-1}|x_t,x_0) = N(x_{t-1};\tilde{\mu}_t(x_t,x_0),\tilde{\beta}_tI),
\end{align}
where $\tilde{\mu}_t(x_t,x_0) := \frac{\sqrt{\bar{\alpha}_{t-1}}(1-\alpha_t)}{1-\bar{\alpha}_t}x_0 +  \frac{\sqrt{\alpha_t}(1-\bar{\alpha}_{t-1})}{1-\bar{\alpha}_t}x_t$, $\tilde{\beta}_t:= \frac{1-\bar{\alpha}_{t-1}}{1-\bar{\alpha}_t}(1-\alpha_t)$, and $\bar{\alpha_t} := {\textstyle \prod_{s=1}^{t}} \alpha_s$.
The full derivations are included in the supplementary material.

Note that the above-defined forward diffusion formulation has no learnable parameters, and the reverse diffusion step cannot be applied due to having no access to $x_0$ in the inference stage. Therefore, we further introduce the learnable reverse denoise process for estimating $x_0$ from $x_T$.

{\bf The reverse denoise process.}
The DDPM is trained to reverse the process in Eq. \ref{1} by learning the denoise network $f_\theta$ in the reverse process. Specifically, the denoise network estimates $f_\theta(x_t,t)$ to replace $x_0$ in Eq. \ref{3}.
Note that $f_\theta(x_t,t)$ directly predicts the Gaussian noise $\varepsilon$, instead of $x_0$. While, the estimated $\varepsilon$ deterministicly corresponds to $\hat{x}_0$ via Eq. \ref{2}.

\begin{equation}
	\small
	\label{4}
	\begin{aligned}
    p_\theta(x_{t-1}|x_t) &= q(x_{t-1}|x_t,f_\theta(x_t,t))\\
    &= N(x_{t-1};\mu _\theta(x_t,t),{\textstyle \sum_{\theta}^{}} (x_t,t)).\\
    \end{aligned}
\end{equation}
\begin{equation}
	\small
	\label{5}
	\mu _\theta(x_t,t) = \tilde{\mu_t}(x_t,x_0),  {\textstyle \sum_{\theta}^{}} (x_t,t) = \tilde{\beta_t}I.
\end{equation}

Similarly, the mean and variance in the reverse Gaussian distribution Eq. \ref{4} can be determined by replacing $x_0$ in $\tilde{\mu}_t(x_t,x_0)$ and $\tilde{\beta}_t$ with the learned $\hat{x}_0$

{\bf Training objective and sampling process.}
As mentioned above, $f_\theta(x_t,t)$ is trained to approach the Gaussian noise $\varepsilon$. Thus the final training objective is:
\begin{align}
\label{6}
    L=E_{t,x_0,\varepsilon}\left \| \varepsilon -f_\theta(x_t,t) \right \|_1.
\end{align}

The sampling process in the inference stage is done by running the reverse process. Starting from a pure Gaussian noise $x_T$, we iteratively apply the reverse denoise transition $p_\theta(x_{t-1}|x_t)$ T times, and finally get the clear output $x_0$.

\section{The CLIP Text Encoder}  \label{sec:encoder}
Causal mask and padding mask are two basic concepts in the transformer framework \cite{vaswani2017attention} of natural language processing (NLP). NLP is a sequential task and the length of the text input is varying. To this end, the causal mask is introduced to enable sequential property. Concretely, the causal mask enables that the word embedding $e_{i}$ is only correlated with word embedding $e_{j} (j \le i)$. To deal with the varying length of the text input, transformers in NLP pad it to a given max text length. Then, padding mask is introduced to ensure that the padded word embedding is meaningless and has 
no information of the input text.

In this paper, we study the text encoder of CLIP \cite{radford2021learning}, which employs the transformer in NLP mode. The CLIP text encoder employs a causal mask and omits the padding mask. This design choice provides the first insight on the context correlation within the text embedding. Specifically, (1) the use of a causal mask ensures that information in a specific word embedding is solely correlated with the word embedding preceding it. (2) The absence of the padding mask endows the padding embedding with information from the semantic embedding.


\begin{figure*}[t]
    \begin{center}
	\includegraphics[width=0.98\linewidth]{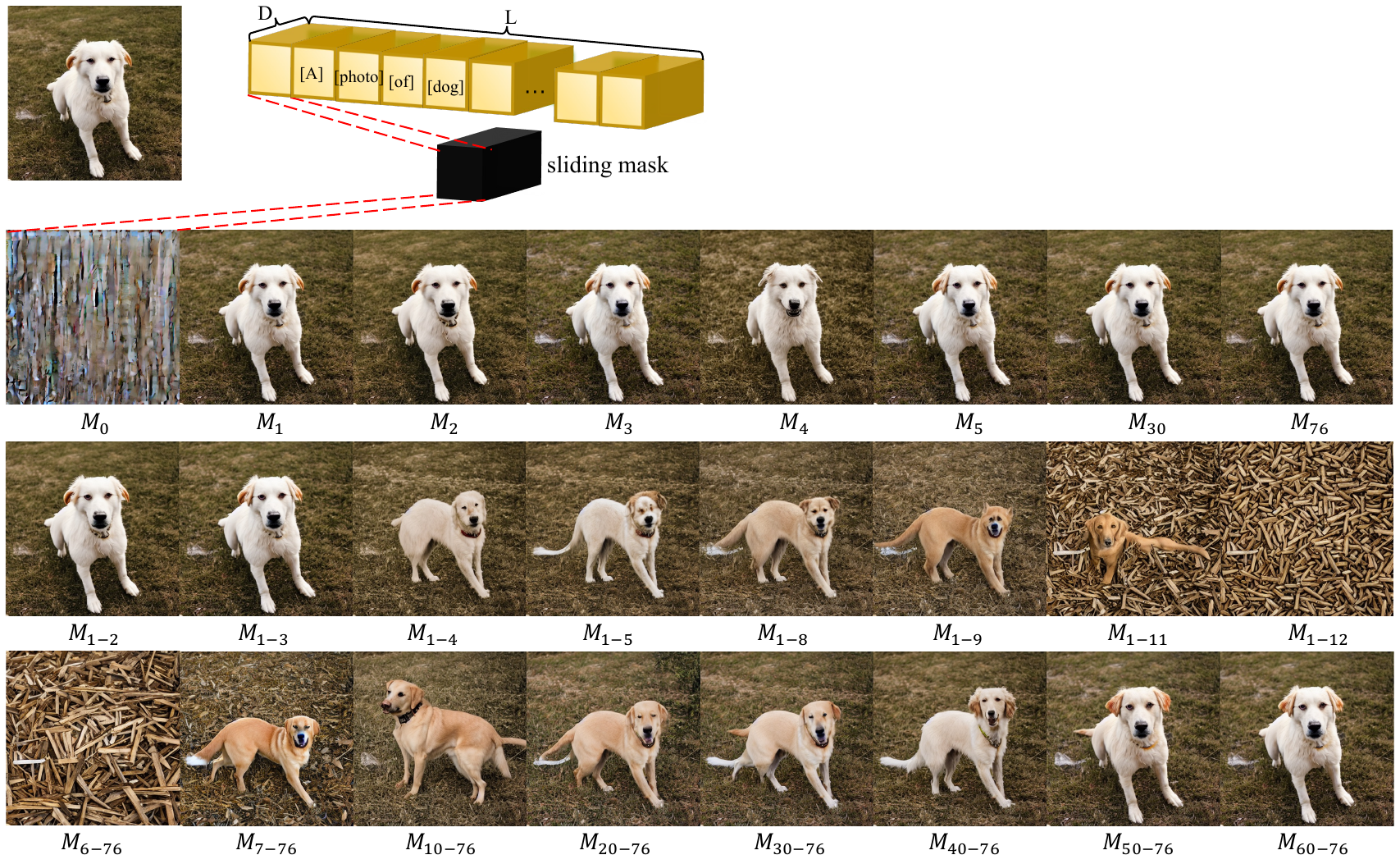}
    \end{center}
    \caption{Examples of the \textit{mask-then-generate} strategy. The text is ``A photo of dog".} 
    \label{fig:mask1}
\end{figure*}

\begin{figure*}[h]
    \begin{center}
	\includegraphics[width=0.98\linewidth]{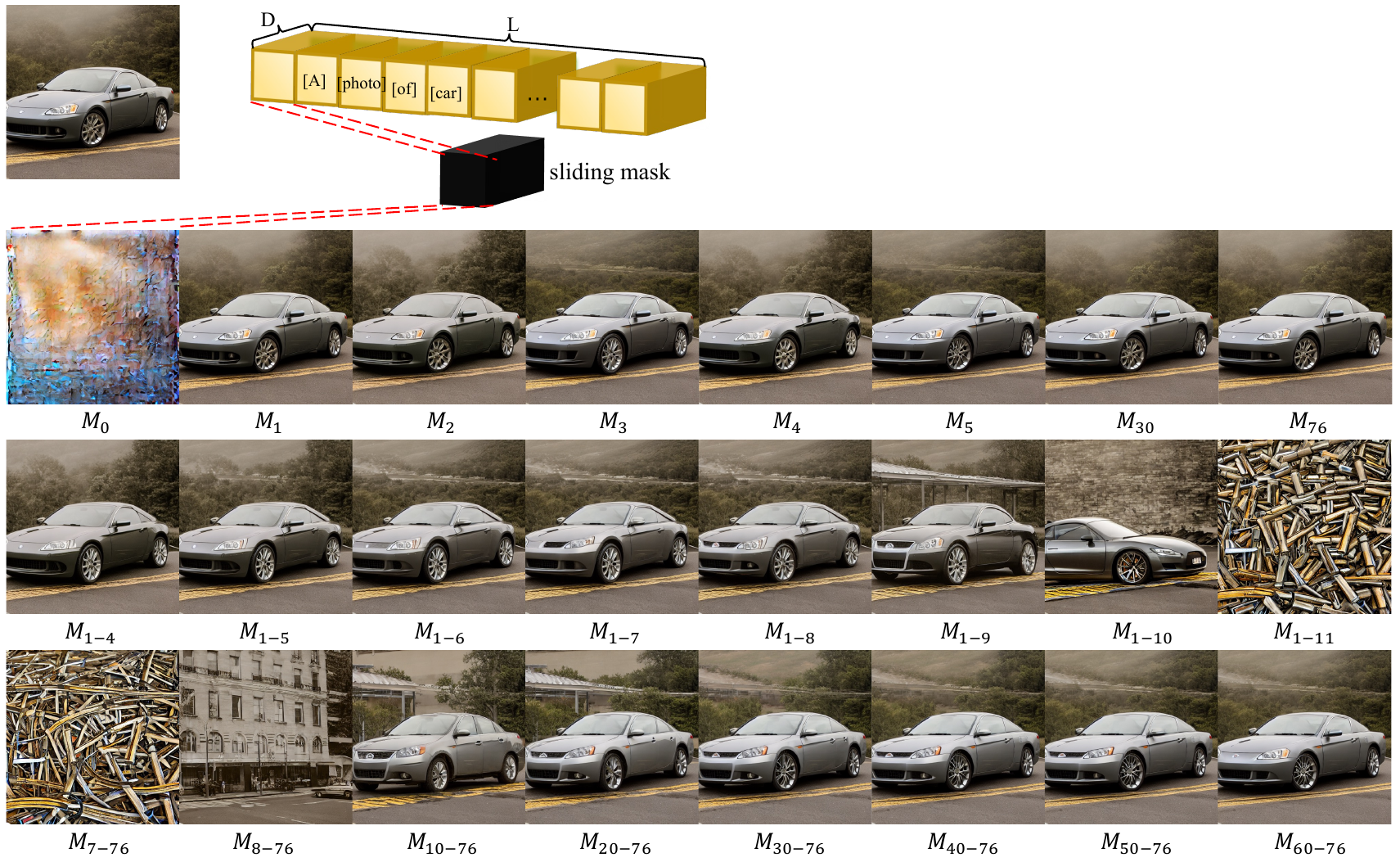}
    \end{center}
    \caption{Examples of the \textit{mask-then-generate} strategy. The text is ``A photo of car".} 
    \label{fig:mask2}
\end{figure*}

\section{\textit{Mask-then-Generate} Strategy}  \label{sec:strategy}
Besides the context correlation within the text embedding, we also aim to figure out the significance and meaning of per word embedding. To this end, we adopt a \textit{mask-then-generate} strategy and present an example in the main manuscript. In this part, we depict more examples of this strategy to validate the generality of the conclusions derived in the main manuscript.

We show two detailed examples under the \textit{mask-then-generate} strategy in Fig.~\ref{fig:mask1} and \ref{fig:mask2}.
In the first row, we mask single word embedding and generate the corresponding images. We discover that the absence of single word embedding won't affect the generation of the image, except the first word embedding, BOS. The BOS embedding is semantically meaningless but indispensable for image generation in stable diffusion. The rational may be that the BOS embedding is same across different text embeddings, and the diffusion network has learned this mode during training. 

In the second row, we mask more word embeddings from front to back and generate the corresponding images. We discover that masking the whole semantic embedding (e.g., $M_{1\text{-}5}$) can still generate the desired object (e.g., dog or car). This reveals that the padding embedding also has the information of the input text. Besides, we find that masking the semantic embedding and first few padding embedding will stop the generation (e.g., $M_{1\text{-}11}$). This supports the importance of the semantic embedding and the first few padding embedding that is close to the semantic embedding. 

In the third row, we mask more word embeddings from bask to front and generate the corresponding images. We discover that masking the majority of the padding embedding (e.g., $M_{40\text{-}76}$) won't stop the generation of the target. This reveals that the padding embedding is relatively less important. 

Apart from the \textit{mask-then-generate} strategy, the image editing operations in the main manuscript also support the conclusion about the significance of word embedding. 


\section{Optimizing-based Paradigm}  \label{sec:optimize}
The optimizing-based paradigm adopts the same framework as \cite{wu2023uncovering}. Differently, we only learn the mixing weight between text embeddings and then employ this mixed text embedding across all timesteps. Our method is required to simultaneously keep the background unchanged and alter the object. We experimentally find that the learning-free text embedding manipulation is a desirable initialization.

\section{PCA and SVD}  \label{sec:PCA}

For a matrix $A \in R^{L \times D}$, its singular value decomposition (SVD) can be expressed as follows:
\begin{equation}
    \small
    \label{eq:loss}
    \begin{aligned}
    A = U \Sigma V^T,\\
    \end{aligned}
\end{equation}
where $\Sigma = diag(\sigma) \in R^{L \times D}$ is the singular value matrix, with the singular values in descending order, $U \in R^{L \times L}$ is the left singular matrix, and $V^T \in R^{D \times D}$ is the right singular matrix.
Further, $V$ is calculated with the eigenvalue decomposition as follows:
\begin{equation}
    \small
    \label{eq:loss}
    \begin{aligned}
    A^T A &= (U \Sigma V^T)^T U \Sigma V^T = V\Sigma^T U^T U \Sigma V^T \\
    &= V\Sigma^T \Sigma V^T = V\Sigma^2 V^T\\
    \end{aligned}
\end{equation}

Besides, in concrete implementation, PCA is also achieved via SVD due to the large computation to calculate the covariance matrix.

\section{More Samples}  \label{sec:sample}
\begin{figure*}[h]
    \begin{center}
	\includegraphics[width=0.98\linewidth]{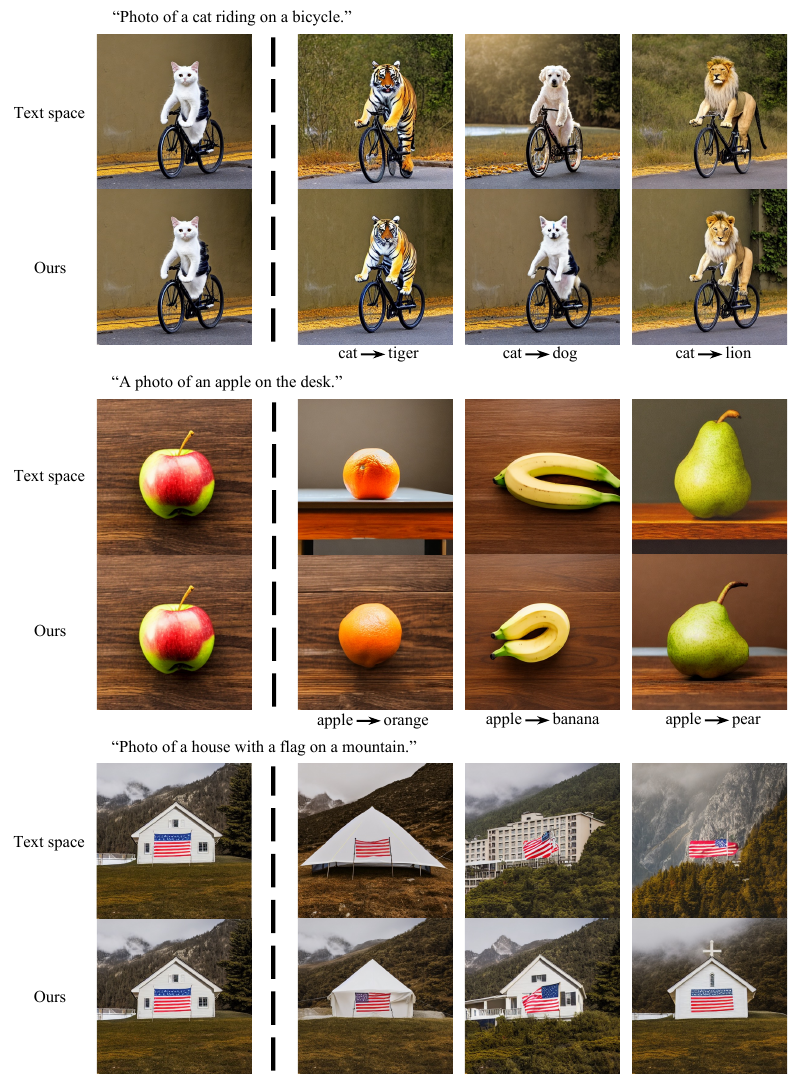}
    \end{center}
    \caption{More examples of object replacement.} 
    \label{fig:edit_rep}
\end{figure*}

\begin{figure*}[h]
    \begin{center}
	\includegraphics[width=0.98\linewidth]{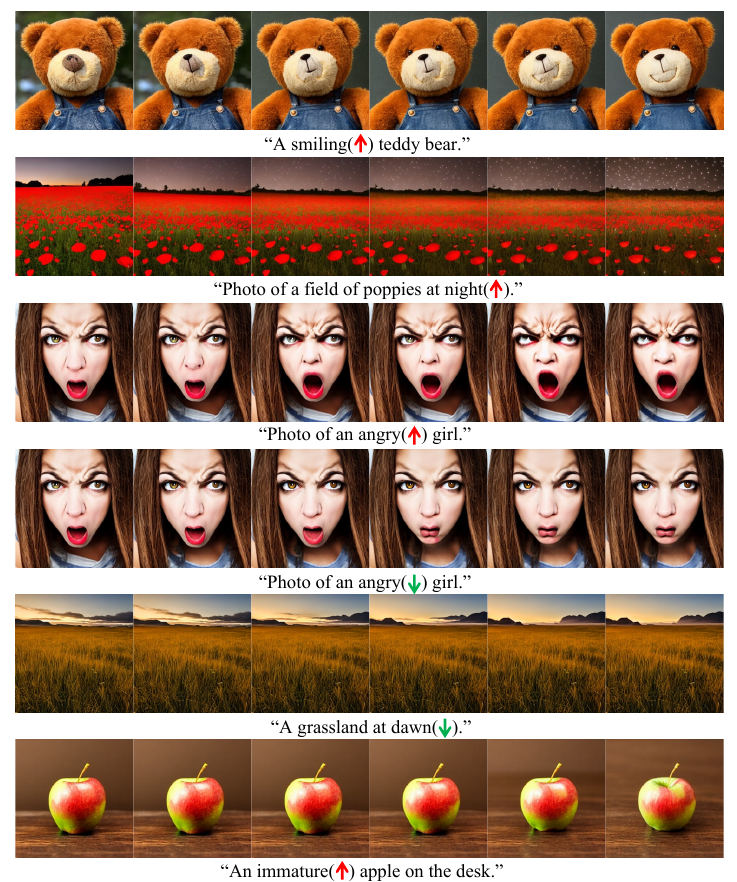}
    \end{center}
    \caption{More examples of fader control.} 
    \label{fig:edit_fader}
\end{figure*}

\begin{figure*}[h]
    \begin{center}
	\includegraphics[width=0.98\linewidth]{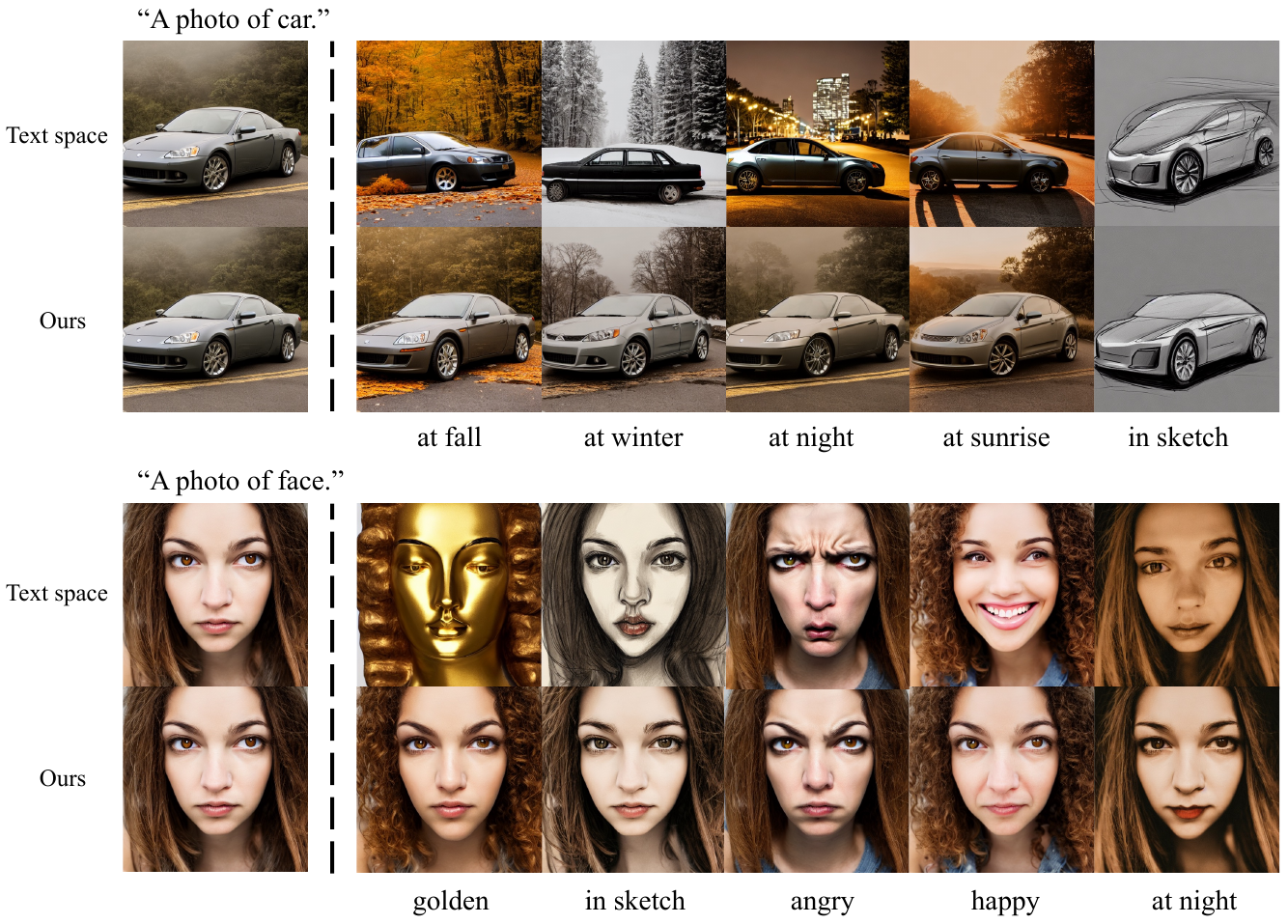}
    \end{center}
    \caption{More examples of style transfer.} 
    \label{fig:edit_style}
\end{figure*}

\begin{figure*}[h]
    \begin{center}
	\includegraphics[width=0.98\linewidth]{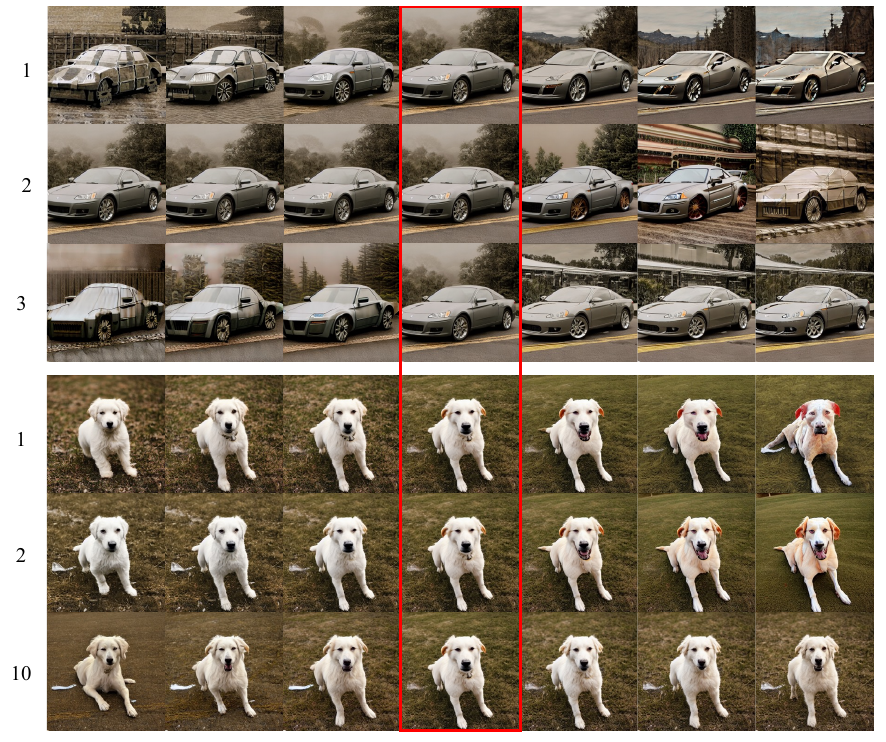}
    \end{center}
    \caption{Examples of semantic directions on the right singular vector. The number on the left denotes the index of the singular vector.} 
    \label{fig:edit_SVDright}
\end{figure*}

\begin{figure*}[h]
    \begin{center}
	\includegraphics[width=0.98\linewidth]{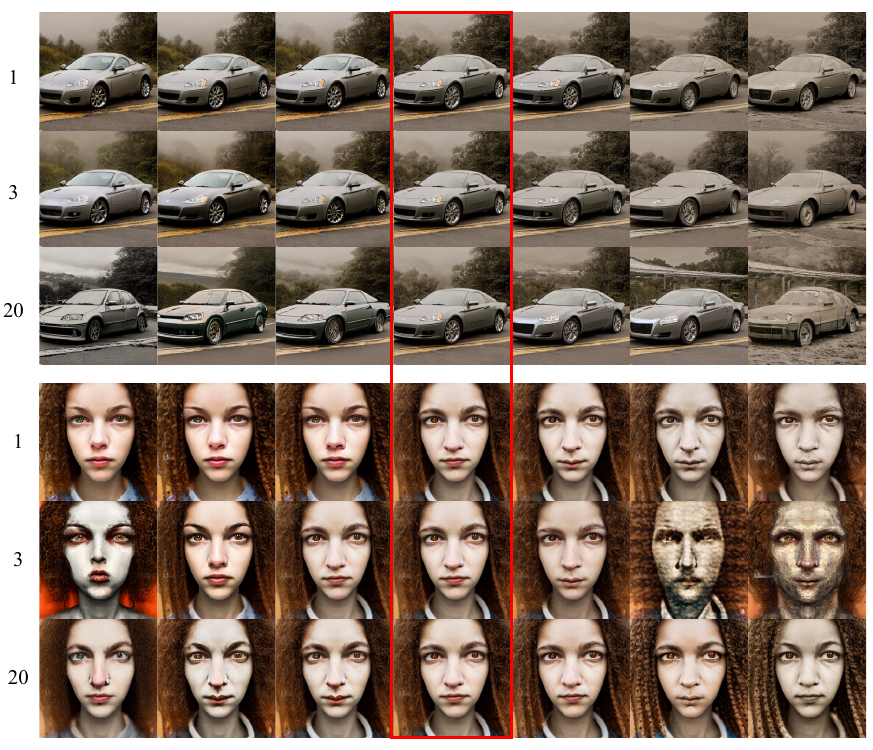}
    \end{center}
    \caption{Examples of semantic directions on the left singular vector. The number on the left denotes the index of the singular vector.} 
    \label{fig:edit_SVDleft}
\end{figure*}

In this part, we show more visual examples to demonstrate the generality of our method. Specifically, we show more samples on object replacement in Fig.~\ref{fig:edit_rep}. More image fader control samples are depicted in Fig.~\ref{fig:edit_fader}. Additional image style transfer samples are depicted in Fig.~\ref{fig:edit_style}.
We also present more samples of semantic directions in Fig.~\ref{fig:edit_SVDright} and Fig.~\ref{fig:edit_SVDleft}.

\newpage
{
    \small
    \bibliographystyle{ieeenat_fullname}
    \bibliography{main}
}